\documentclass[runningheads]{llncs}
\pdfoutput=1
\usepackage{times}
\usepackage{latexsym}
\usepackage{amsmath}
\usepackage{url}
\usepackage{amssymb}
\usepackage{amsfonts}
\usepackage{graphicx}
\usepackage{tabularx}
\usepackage{multirow}
\usepackage{mathtools,nccmath}
\usepackage{arydshln}
\usepackage{hyperref} 
\hypersetup{hidelinks}
\usepackage[usenames,dvipsnames]{color}

\title{End-to-end neural relation extraction using deep biaffine attention}

\author{Dat Quoc Nguyen \and Karin Verspoor}
\institute{The University of Melbourne, Australia\\
\email{\{dqnguyen, karin.verspoor\}@unimelb.edu.au}
}

\begin{document}
\maketitle

\begin{abstract}
  We propose a neural network model for joint extraction of named entities and relations between them, without any hand-crafted features. The key contribution of our model is to extend a  BiLSTM-CRF-based entity recognition model with a deep biaffine attention layer to model second-order interactions between latent features for relation classification, specifically attending to the role of an entity in a directional relationship. 
  On the benchmark ``relation and entity recognition"  dataset CoNLL04, experimental results show that our model outperforms previous models, producing new state-of-the-art performances.   
\end{abstract}

\section{Introduction}

Extracting entities and their semantic relations from raw text is a key information extraction task.  For example, given the sentence ``\colorbox{SkyBlue}{David Foster} is the \colorbox{Salmon}{AP} 's \colorbox{Yellow}{Northwest} regional reporter , based in \colorbox{Yellow}{Seattle}'' in the CoNLL04 dataset \cite{roth-yih:2004:CONLL}, our goal is to recognize ``David Foster'' as person, ``AP'' as  organization, and ``Northwest'' and ``Seattle'' as location entities, then classifiy entity pairs to extract structured information: \textsl{Work\_For}(David Foster, AP), \textsl{OrgBased\_In}(AP, Northwest) and \textsl{OrgBased\_In}(AP, Seattle). Such information is  useful  in many other NLP tasks. Especially in  IR  applications such as  entity search, structured search and question answering, it helps provide end users with significantly better search experience   \cite{Jiang2012,gks563,Blanco:2013}.


A common relation extraction approach  is to construct pipeline systems  with separate sub-systems for the two tasks of  named entity recognition and relation classification \cite{bach2007}. More recently, end-to-end systems which jointly learn to extract entities and relations have been proposed with strong potential to obtain  high performance  \cite{RothYi07}. 
Traditional joint approaches  are feature-based supervised learning methods which employ numerous  syntactic and lexical  features based on external NLP tools as well as knowledge base resources  \cite{Kate:2010,miwa-sasaki:2014:EMNLP2014,li-ji:2014:P14-1}. 

State-of-the-art relation extraction performance has been obtained by end-to-end models based on neural networks. Specifically, Gupta et al. (2016)  \cite{gupta-schutze-andrassy:2016:COLING} proposed a RNN-based  model which achieved top results on the  CoNLL04 dataset. Their approach 
relies on various manually extracted features. Other neural models  employ  dependency parsing-based information \cite{miwa-bansal:2016:P16-1,pawarEACLlong,zhang-zhang-fu:2017:EMNLP2017}.  
 In particular, Miwa and Bansal (2016) \cite{miwa-bansal:2016:P16-1} applied bottom-up and top-down tree-structured LSTMs to model dependency paths between entities. Zhang et al. (2017) \cite{zhang-zhang-fu:2017:EMNLP2017} integrated implicit syntactic information by using latent feature representations extracted from a pre-trained BiLSTM-based dependency parser.  Zheng et al. (2017) \cite{ZhengHLBXHX17} used a softmax layer on top of a BiLSTM for entity recognition, and a CNN on top of the  BiLSTM for classifying  relations  \cite{Nguyen2016}.   Adel and Sch\"utze (2017) \cite{adel-schutze:2017:EMNLP2017} assumed that entity boundaries are given, and   trained a CNN to  extract context  features around the entities, and using these features  for entity and relation classification.  
Recently, Wang  et al. (2018) \cite{ijcai2018-620} formulated the joint entity and relation extraction problem as a directed graph and proposed a BiLSTM- and transition-based approach to generate the graph incrementally. Bekoulis et al. (2018) \cite{Bekoulis18emnlp}   extended the multi-head selection-based   joint model \cite{BEKOULIS201834} with adversarial training. 
In \cite{BEKOULIS201834,zheng-EtAl:2017:Long,katiyar-cardie:2017:Long}, the joint  task is  formulated as a sequence tagging problem, and  a BiLSTM with a softmax output layer can then be used for joint prediction.

\begin{figure}[!t]
\centering
\includegraphics[width=9.5cm]{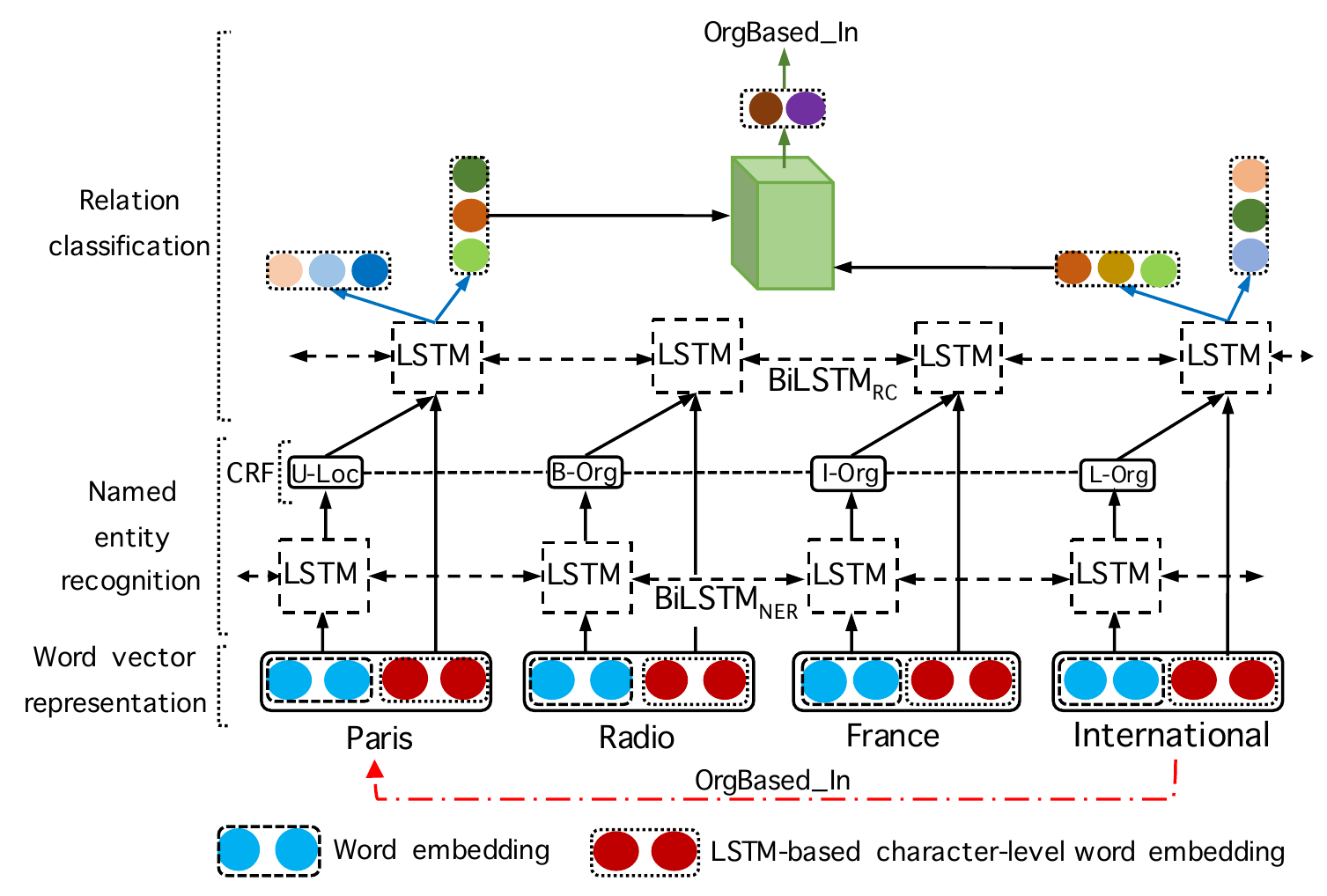}
\vspace{-10pt}
\caption{Illustration of our  model. Linear transformations are not shown   for simplification.}
\label{fig:architecture}
\vspace{-5pt}
\end{figure}

In this paper, we present a novel end-to-end neural  model for joint entity and relation extraction. As illustrated in Figure \ref{fig:architecture}, our model architecture can be viewed as a mixture of a named entity recognition (NER) component  and a relation classification  (RC) component. Our NER component employs a  BiLSTM-CRF architecture  \cite{HuangXY15} to predict entities from input word tokens. Based on both the input words  and the predicted NER labels, the RC component uses another BiLSTM to learn latent features relevant for relation classification. 
In most previous neural joint models, the  relation classification part relies on a common ``linear'' concatenation-based  mechanism over the latent features associated with entity  pairs, i.e. the latent features are first concatenated into a single feature vector which is then linearly transformed before being fed into a softmax classifier.   In contrast, our RC component takes  into account second-order interactions over the latent features via a tensor. In particular,  for relation classification we propose a novel use of  the  deep \textit{biaffine} attention mechanism \cite{DozatM17}  which was first introduced in  dependency parsing.

Experimental results on the benchmark ``relation and entity recognition"   dataset CoNLL04 \cite{roth-yih:2004:CONLL} show that our model outperforms previous models, obtaining new state-of-the-art scores. In addition, 
using  the {biaffine} attention improves  the performance compared to using the linear mechanism significantly. We also provide an  ablation study to investigate effects of different contributing factors in our  model.

\section{Our proposed model}

This section details  our end-to-end relation extraction model. 
Given an input  sequence of $n$ word tokens $w_1$, $w_2$, ..., $w_n$, we use a vector $\mathbf{v}_i$ to represent each $i^{th}$ word $w_i$ by concatenating word embedding $\mathbf{e}^{(\textsc{w})}_{w_i}$ and character-level word embedding $\mathbf{e}^{(\textsc{c})}_{w_i}$:

 \begin{equation}
\mathbf{v}_{i} =  \mathbf{e}^{(\textsc{w})}_{w_i} \circ \mathbf{e}^{(\textsc{c})}_{w_i} 
\end{equation} 

Here, for each word type $w$, we use a one-layer BiLSTM ($\mathrm{BiLSTM}_{\text{char}}$)  to learn its  character-level word embedding $\mathbf{e}^{(\textsc{c})}_{w}$ \cite{ballesteros-dyer-smith:2015:EMNLP}. 

\medskip

\noindent\textbf{Named entity recognition (NER):} The NER component feeds the  sequence of vectors $\mathbf{v}_{1:n}$ with an additional context position index $i$ into another BiLSTM ($\mathrm{BiLSTM}_{\text{NER}}$)  to learn a ``latent'' feature vector  representing the $i^{th}$ word token.
Then the  NER component performs linear transformation of  each latent feature vector  by using a single-layer feed-forward  network ($\mathrm{FFNN}_{\text{NER}}$):

\begin{equation}
\mathbf{h}_{i} =  \mathrm{FFNN}_{\text{NER}}\big(\mathrm{BiLSTM}_{\text{NER}}(\mathbf{v}_{1:n}, i)\big)
\end{equation} 

\noindent The output layer size of $\mathrm{FFNN}_{\text{NER}}$ is the number of BIOLU-based NER labels \cite{ratinov-roth:2009:CoNLL}. 
The  NER component feeds the output vectors $\mathbf{h}_{1:n}$  into a linear-chain CRF layer \cite{Lafferty:2001} for NER label  prediction. A cross-entropy loss $\mathcal{L}_{\text{NER}}$ is computed during training, while the Viterbi algorithm is used for decoding. Our NER component thus is the BiLSTM-CRF model  \cite{HuangXY15} with additional LSTM-based character-level word embeddings \cite{lample-EtAl:2016:N16-1}.

\medskip

\noindent\textbf{Relation classification (RC):}   Assume that $t_1, t_2, ..., t_n$ are   NER labels predicted by the NER component for the input words.
We represent each $i^{th}$ predicted label by a vector embedding $\mathbf{e}_{t_i}$. We  create a sequence of vectors $\boldsymbol{x}_{1:n}$ in which each $\boldsymbol{x}_{i}$ is computed as:

\begin{equation}
\mathbf{x}_{i} = \mathbf{e}_{t_i}  \circ   \mathbf{v}_{i} 
\end{equation}

\noindent As for NER, the RC component also uses a BiLSTM ($\mathrm{BiLSTM}_{\text{RC}}$)  to learn another set of latent feature vectors, but from the  sequence $\boldsymbol{x}_{1:n}$:

\begin{equation}
\mathbf{r}_{i} = \mathrm{BiLSTM}_{\text{RC}}(\boldsymbol{x}_{1:n}, i)
\end{equation}

\noindent  The  RC component  further  uses these latent vectors  $\mathbf{r}_{i}$ for relation classification. 

We propose a novel use of the  deep \textit{biaffine} attention mechanism \cite{DozatM17} for relation classification.  The {biaffine} attention mechanism was proposed for dependency parsing \cite{DozatM17}, helping to produce the best reported   parsing  performance to date \cite{dozat-qi-manning:2017:K17-3}. First, to encode the directionality of a relation, we use two single-layer feed-forward networks to project each  $\mathbf{r}_{i}$ into \textit{head} and \textit{tail} vector representations which correspond to whether the $i^{th}$ word serves as the head or tail argument of the relation:

\begin{eqnarray}
\mathbf{h}_{i}^{(\text{head})} &= &\mathrm{FFNN}_{\text{head}}(\boldsymbol{r}_{i}) \\
\mathbf{h}_{i}^{(\text{tail})} &= &\mathrm{FFNN}_{\text{tail}}(\boldsymbol{r}_{i})
\end{eqnarray}

Following \cite{miwa-bansal:2016:P16-1}, our RC component incrementally constructs relation candidates using all possible combinations of the last word tokens of predicted entities, i.e. words with L or U labels. We assign an entity pair to a negative relation class (NEG) when the  pair has no relation or when the predicted entities are not correct. For example, for  Figure \ref{fig:architecture}, we would have two relation candidates: \textsl{NEG}(Paris, International) and \textsl{OrgBased\_In}(International, Paris). 
  Then for each head-tail candidate pair ($w_j, w_k$), we  apply the biaffine attention operator:

\begin{eqnarray}
  \mathbf{s}_{j,k}  &=&\mathrm{Biaffine}\Big(\mathbf{h}_{j}^{(\text{head})}, \mathbf{h}_{k}^{(\text{tail})}\Big) \\
\mathrm{Biaffine}\big(\mathbf{y}_{1}, \mathbf{y}_{2}\big) &=& \underbrace{\mathbf{y}_{1}^{\mathsf{T}} \mathbf{U} \mathbf{y}_{2}}_{\mathrm{Bilinear}}  +  \underbrace{\mathbf{W}(\mathbf{y}_{1} \circ  \mathbf{y}_{2}) +  \mathbf{b}}_{\mathrm{Linear}}
\end{eqnarray}


\noindent   where $\mathbf{U}$, $\mathbf{W}$, $\mathbf{b}$ are a $m \times l  \times m$ tensor, a  $l \times (2 * m)$ matrix and a bias vector, respectively. Here, $m$ is the size of the output layers of both $\mathrm{FFNN}_{\text{head}}$ and $\mathrm{FFNN}_{\text{tail}}$, while  $l$ is the number of relation classes (including NEG).   
Next, the RC component  feeds the output vectors $\mathbf{s}_{j,k}$ of the biaffine attention layer into a softmax layer for relation prediction. Another cross-entropy loss $\mathcal{L}_{\text{RC}}$ is then computed during training.

\medskip

\noindent\textbf{Joint learning:} The objective loss of our joint model is the sum of the NER and RC losses:    $\mathcal{L} = \mathcal{L}_{\text{NER}} + \mathcal{L}_{\text{RC}}$. Model parameters are then learned to minimize $\mathcal{L}$.


\section{Experiments}

\subsection{Experimental setup}

\noindent\textbf{Evaluation scenarios:} We evaluate our joint model on two  evaluation setup scenarios: \textbf{(1)} \underline{NER\&RC}: A realistic  scenario where entity boundaries are \textit{not} given. \textbf{(2)} \underline{EC\&RC}: A less realistic scenario where the  entity boundaries are given \cite{RothYi07,Kate:2010,miwa-sasaki:2014:EMNLP2014}. Thus  the NER task which identifies both entity boundaries and classes reduces to the \textit{entity classification} (\textbf{EC}) task.  Following \cite{miwa-sasaki:2014:EMNLP2014}, we  encode the  gold entity boundaries  in the BILOU scheme. Then we represent each B, I, O, L or U boundary tag as  a vector embedding. As a result,  the  vector $\mathbf{v}_i$  in Equation 1 now also includes the  boundary tag embedding in addition to the word embedding and character-level word embedding.

\medskip

\noindent\textbf{Dataset:} We use the benchmark   ``entity and relation recognition'' dataset  CoNLL04 from \cite{roth-yih:2004:CONLL}. Following \cite{Bekoulis18emnlp,BEKOULIS201834}, we use the 64\%/16\%/20\% training/development/test pre-split available from Adel and Sch\"utze (2017) \cite{adel-schutze:2017:EMNLP2017}, in which the test set was previously also used by Gupta et al. (2016) \cite{gupta-schutze-andrassy:2016:COLING}. 

\medskip

\noindent\textbf{Implementation:} Our model is implemented using  \textsc{DyNet} v2.0 \cite{dynet}. We optimize the objective loss using Adam  \cite{KingmaB14}, no mini-batches and run for 100 epochs.  We  compute the average of NER/EC score and RC score after each training epoch. We choose the model with the highest average score on the development set, which is then applied to the test set for the final evaluation phase. More details of the implementation as well as optimal hyper-parameters are  in the Appendix. Our code is available at: \url{https://github.com/datquocnguyen/jointRE}  

\medskip

\noindent\textbf{Metric:} Similar to  previous works in Table  \ref{tab:comparison}, we use the macro-averaged F1-score over the entity classes to score NER/EC and over the relation classes to score RC.  
More details of the metric are also in the Appendix. 
 Unlike previous neural models, we report   results as mean and standard deviation of the scores over 10 runs with 10 random seeds.

\begin{table}[!t]
\centering
\setlength{\tabcolsep}{0.5em}
\caption{Comparison with the previous state-of-the-art results on the \textbf{test} set. Recall that Setup 2 uses gold entity boundaries while Setup 1 does not. The subscript denotes the standard deviation.   \textbf{(F)} refers to the use of extra feature types such as POS tag-based or  dependency parsing-based features. 
 Although using the same test set,  Gupta et al. (2016)  \cite{gupta-schutze-andrassy:2016:COLING} reported results on a 80/0/20 training/development/test split rather than our {64/16/20} split. Results in the last two rows are  just for reference, not for comparison, due to a random sampling of the  test set. In particular,  Miwa and Sasaki (2014)   \cite{miwa-sasaki:2014:EMNLP2014}  used the 80/0/20 split for Setup 1  and performed 5-fold cross validation (i.e. sort of equivalent to 80/0/20) for Setup 2, while Zhang et al. (2017)   \cite{zhang-zhang-fu:2017:EMNLP2017}  used a 72/8/20 split. }
\begin{tabular}{l|ll|ll}
\hline
\multirow{2}{*}{\bf Model} & \multicolumn{2}{c|}{\bf Setup 1} & \multicolumn{2}{c}{\bf Setup 2}\\
\cline{2-5}
& NER & RC & EC & RC \\
\hline
Gupta et al. (2016) \cite{gupta-schutze-andrassy:2016:COLING} & \_ & \_ &  88.8 & 58.3  \\
Gupta et al. (2016) \cite{gupta-schutze-andrassy:2016:COLING} \textbf{(F)} & \_ & \_ &  92.4 & 69.9  \\
Adel and Sch\"utze (2017) \cite{adel-schutze:2017:EMNLP2017}& \_ & \_ & 82.1 & 62.5\\
Bekoulis et al. (2018) \cite{Bekoulis18emnlp} & 83.6 & 62.0 & 93.0 & 68.0 \\
Bekoulis et al. (2018)  \cite{BEKOULIS201834} & 83.9 & 62.0& 93.3 & 67.0 \\
\hline
Our joint model & \textbf{86.2}\textsubscript{0.5} & \textbf{64.4}\textsubscript{0.6}& \textbf{93.8}\textsubscript{0.4} & \textbf{69.6}\textsubscript{0.7} \\
\hline
Miwa and Sasaki (2014) \cite{miwa-sasaki:2014:EMNLP2014} \textbf{(F)}  & 80.7 & 61.0 & 92.3 & 71.0\\
Zhang et al. (2017) \cite{zhang-zhang-fu:2017:EMNLP2017} \textbf{(F)}& 85.6 & 67.8 & \_ & \_  \\
\hline
\end{tabular}
\label{tab:comparison}
\end{table}

\subsection{Main results}

\textbf{End-to-end results:} The first six rows in Table \ref{tab:comparison} compare our  results with previous  state-of-the-art published results on the same test set.  In particular, our model obtains 2+\% absolute higher NER and RC scores (Setup 1) than the BiLSTM-CRF-based multi-head selection  model \cite{BEKOULIS201834}. We also obtain 7+\% higher EC and RC scores (Setup 2) than Adel and Sch\"utze (2017) \cite{adel-schutze:2017:EMNLP2017}.  Note that Gupta et al. (2016)  \cite{gupta-schutze-andrassy:2016:COLING}  use the same test set as we do, however they report final results on a 80/0/20 training/development/test split rather than our {64/16/20}, i.e.  Gupta et al. (2016)  use a larger training set, but producing about 1.5\% lower EC score and similar RC score against ours.  These  results show that our model performs better than previous state-of-the-art  models, using the same  setup.

In Table \ref{tab:comparison}, the last two rows  present results reported in \cite{miwa-sasaki:2014:EMNLP2014} and \cite{zhang-zhang-fu:2017:EMNLP2017} on the  dataset CoNLL04. However, these results are \textit{not} comparable due to their random sampling of the test set, i.e. using different train-test splits.   
Both Miwa and Sasaki (2014)  \cite{miwa-sasaki:2014:EMNLP2014} 
and Zhang et al. (2017) \cite{zhang-zhang-fu:2017:EMNLP2017} 
employ additional extra features based on external NLP tools and use larger training sets than ours. Specifically, Zhang et al. (2017) 
 integrate syntactic features by using a pre-trained BiLSTM-based  dependency parser to extract BiLSTM-based latent feature representations for words in the input  sentence,  and then using these latent representations directly as part of the input embeddings in their model. We plan to extend our model with their  syntactic integration approach to further improve our model performance in future work.

 \begin{table}[!t]
\centering
\caption{Ablation results on the \textbf{development} set.  * and ** denote the statistically significant differences against the  \underline{full} results at $p < 0.05$ and $p < 0.01$, respectively (using the  two-tailed paired t-test). (a) Without using the character-level word embeddings. (b) Using a softmax layer for NER label prediction instead of the CRF layer. (c)  Without using the NER label embeddings in our RC component, i.e. Equation 3 would become $\mathbf{x}_{i} = \mathbf{v}_{i}$. (d)  Without using the $\mathrm{Bilinear}$ part in Equation 8, i.e., $\mathrm{Biaffine}$ would be a common $\mathrm{Linear}$ mechanism.  (e)  Without using the $\mathrm{Linear}$ part in Equation 8, i.e., $\mathrm{Biaffine}$ reduces to $\mathrm{Bilinear}$.}
\setlength{\tabcolsep}{0.5em}
\begin{tabular}{l|ll|ll}
\hline
\multirow{2}{*}{\bf Model} & \multicolumn{2}{c|}{\bf Setup 1} & \multicolumn{2}{c}{\bf Setup 2}\\
\cline{2-5}
& NER & RC & EC & RC \\
\hline
Pipeline & 87.3\textsubscript{0.6}	 & 66.3\textsubscript{0.8} & 93.4\textsubscript{0.6}	& 72.9\textsubscript{0.6} \\
\hline
Joint model (full) & 87.1\textsubscript{0.5}	& 66.9\textsubscript{0.8} & 93.3\textsubscript{0.5}	& 73.3\textsubscript{0.6}	 \\
\hdashline
\ \ (a) w/o Character & 82.7$_{0.5}^{**}$	& 63.0$_{0.7}^{**}$	& 93.1\textsubscript{0.6}	& 73.4\textsubscript{0.8}	  \\
\ \ (b)  w/o CRF &  86.4$_{0.5}^{*}$	 &  	66.0$_{0.8}^{*}$ & 93.5\textsubscript{0.4}	& 73.2\textsubscript{0.6}	 \\ 
\ \ (c)  w/o Entity & 87.1\textsubscript{0.5} & 64.7$_{0.9}^{**}$	& 93.3\textsubscript{0.6} 	& 72.1$_{0.7}^{**}$  \\
\ \ (d)  w/o Bilinear 	&  86.6\textsubscript{0.5} 	& 65.4$_{0.7}^{**}$ & 93.4\textsubscript{0.5} 	& 72.0$_{0.7}^{**}$  \\
\ \ (e)  w/o Linear	& 86.8\textsubscript{0.6} 	& 65.9$_{0.7}^{*}$ & 93.3\textsubscript{0.5} 	& 72.6$_{0.5}^{*}$ \\
\hline
\end{tabular}
\label{tab:ablation}
\end{table}

\medskip

\noindent \textbf{Ablation analysis:} 
We  provide in Table  \ref{tab:ablation} the results of a pipeline approach where we treat our two NER and RC  components  as independent networks, and  train them separately.  Here, the  RC network uses gold NER labels when training, and uses predicted labels produced by the NER network when decoding.  We find that the   joint approach does slightly better   than the pipeline approach in relation classification,  although the differences  are not significant. A similar observation is also found in \cite{miwa-bansal:2016:P16-1}. Also, in preliminary experiments, we  do not find any significant difference  in performance of our joint  model when feeding gold NER labels instead of predicted NER labels into the RC component during training. This is not surprising as the training NER score is at 99+\%. 

Table \ref{tab:ablation} also presents ablation tests over 5  factors of our joint  model on the development set. In particular,  Setup 1 performances significantly degrade by 4+\% absolutely, when not using the character-level word embeddings. 
The performances also decrease when using a softmax classifier for NER label prediction rather than a CRF layer (here, the decrease is  significant). In contrast, we do not find any significant difference in Setup 2 scores when not using either the character-level  embeddings or the CRF layer, clearly showing the usefulness of the given gold  entity boundaries. 
The 3 remaining  factors, including removing NER label embeddings and  not taking either the $\mathrm{Bilinear}$ or $\mathrm{Linear}$  part (in Equation 8) into the $\mathrm{Biaffine}$ attention layer,   do not  affect the NER/EC score. However, they significantly decrease the RC score. This  is reasonable because those 3 factors are part of the RC component only, thus  helpful in predicting relations. 
More specifically, using the $\mathrm{Biaffine}$ attention produces about 1.5\%  significant improvements to a common $\mathrm{Linear}$ transformation mechanism in relation classification, i.e., ``w/o Bilinear'' results  against the full results in Table \ref{tab:ablation}: 65.4\% vs. 66.9\%  and 72.0\%   vs. 73.3\% (although using $\mathrm{Biaffine}$ increases training time over using $\mathrm{Linear}$ by 35\%, relatively). 

\section{Conclusion}

In this paper, we have presented an end-to-end neural network-based relation extraction model.  Our model employs a BiLSTM-CRF architecture for entity recognition and a biaffine attention mechanism for relation classification. On the benchmark CoNLL04 dataset, our model produces new  state-of-the-art performance.

\medskip

\noindent \textbf{Acknowledgments:} 
This work was supported by the ARC projects DP150101550 and  LP160101469. 

\bibliography{Refs}
\bibliographystyle{splncs04}

\section*{Appendix}

\noindent\textbf{Implementation details:} 
We  apply dropout \cite{JMLR:v15:srivastava14a} with a 67\% keep probability to the inputs of  BiLSTMs and FFNNs. Following  \cite{TACL885},  we also use \textit{word dropout} to learn  an embedding for unknown words: we replace each word token $w$ appearing $\#(w)$ times in the training set  with a special ``unk'' symbol with probability $\mathsf{p}_{unk}(w) = \frac{0.25}{0.25 + \#(w)}$. 

 Word embeddings are initialized by the 100-dimensional pre-trained GloVe word vectors  \cite{pennington-socher-manning:2014:EMNLP2014}, while character and NER label embeddings are  initialized randomly. All these embeddings are then updated during training. 
For learning character-level word embeddings, we   set the size of LSTM hidden states  in $\mathrm{BiLSTM}_{\text{char}}$ to be equal to the size of character embeddings.   
Here, we perform a minimal grid search of hyper-parameters for Setup 1, resulting in the Adam initial learning rate of 0.0005,  the character embedding size of 25, the NER label embedding size of 100,  the size of the output layers of both $\mathrm{FFNN}_{\text{head}}$ and $\mathrm{FFNN}_{\text{tail}}$ at 100,  the number of $\mathrm{BiLSTM}_{\text{NER}}$ and $\mathrm{BiLSTM}_{\text{RC}}$ layers at 2 and the size of LSTM hidden states  in each layer  at 100.  These optimal hyper-parameters for Setup 1 are then reused for Setup 2 where we additionally use  the boundary tag embedding size of 100. 

\medskip

\noindent\textbf{Metric:}  Similar to the previous works, when computing the macro-averaged F1 scores, we omit the entity label ``Other'' and the negative relation  ``NEG''. Here, for NER an entity is predicted correctly if both the entity boundaries and the entity type are correct, while for EC a multi-token entity is considered as correct if at least one of its comprising tokens is predicted correctly. In all cases, a relation is scored as correct if both the argument entities and the relation type are correct.

\end{document}